%% file: arxiv26-representation-fidelity-frame.tex
\documentclass[manuscript,screen]{acmart}

\usepackage{booktabs} 

\newcommand{\Ni}{(1)~}
\newcommand{\Nii}{(2)~}
\newcommand{\Niii}{(3)~}
\newcommand{\Niv}{(4)~}

\usepackage{algorithm2e}
\usepackage{enumitem}
\usepackage{float}
\usepackage{makecell}
\usepackage{lscape}
\usepackage{tikz}
\usepackage{microtype}
\usetikzlibrary{positioning, calc}

\DeclareGraphicsExtensions{.pdf,.ai,.jpg,.png}
\setkeys{Gin}{pagebox=artbox}
\graphicspath{{arxiv26-representation-fidelity-figures/}}

\RequirePackage{type1cm}
\RequirePackage{color}
\RequirePackage{soul}
\setstcolor{blue}
\definecolor{lightgray}{rgb}{0.95,0.95,0.95}
\definecolor{lightgreen}{rgb}{0.56,0.93,0.56}
\definecolor{lightblue}{rgb}{0.3,0.3,0.9}
\definecolor{tgray}{rgb}{0.5,0.5,0.5}

\newsavebox\bscombox
\newcommand{\bscom}[3][]{%
	\sbox{\bscombox}{\fontsize{8}{9}\selectfont#1#2#3}
	\noindent
	\st{#2}{\selectfont
		\color{blue}#3\ifx\\#1\\\else{\fontsize{8}{9}\selectfont\color{violet}[#1]}\fi
	}
}

\author{Theresa Elstner}
\affiliation{
  \institution{Kassel University}
  \country{Germany}
}
\email{theresa.elstner@uni-kassel.de}

\author{Martin Potthast}
\affiliation{
  \institution{Kassel University}
  \country{Germany}
}
\email{martin.potthast@uni-kassel.de}

\begin{document}

\input{arxiv26-representation-fidelity-pre.tex}
\input{arxiv26-representation-fidelity-part1.tex}
\input{arxiv26-representation-fidelity-part2.tex}
\input{arxiv26-representation-fidelity-part3.tex}
\input{arxiv26-representation-fidelity-part4.tex}
\input{arxiv26-representation-fidelity-part5.tex}

\input{arxiv26-representation-fidelity-sum.tex}

\bibliographystyle{ACM-Reference-Format}
\bibliography{arxiv26-representation-fidelity-lit.bib}
\newpage
\input{arxiv26-representation-fidelity-appendix.tex}
\end{document}

%% file: arxiv26-representation-fidelity-pre.tex
\title{Representation Fidelity: Auditing Algorithmic Decisions about Humans Using Self-Descriptions}

\begin{abstract}
This paper introduces a new dimension for validating algorithmic decisions about humans by measuring the fidelity of their representations. \emph{Representation Fidelity} measures if decisions about a person rest on reasonable grounds. 
We propose to operationalize this notion by measuring the distance between two representations of the same person:
\Ni
an externally prescribed input representation on which the decision is based, and
\Nii
a self-description provided by the human subject of the decision, used solely to validate the input representation.
We examine the nature of discrepancies between these representations, how such discrepancies can be quantified, and derive a generic typology of representation mismatches that determine the degree of representation fidelity. We further present the first benchmark for evaluating representation fidelity based on a dataset of loan-granting decisions. Our \texttt{Loan-Granting Self-Representations~Corpus~2025} consists of a large corpus of 30\,000~synthetic natural language self-descriptions derived from corresponding representations of applicants in the German Credit Dataset, along with expert annotations of representation mismatches between each pair of representations.
\end{abstract}

\ccsdesc[500]{Social and professional topics~User characteristics}
\ccsdesc[300]{General and reference~Validation}
\ccsdesc[500]{Computing methodologies~Natural language processing}

\keywords{User Profiles, Algorithmic Decision Making, Explainability}

\maketitle

%% file: arxiv26-representation-fidelity-part1.tex
\section{Introduction}


The design of feature representations for humans is a fundamental step in algorithmic decision-making about them. Guided by the optimization of a decision cost function and trained on past decisions, system designers typically select among the features available to them a subset they deem pertinent to the decision task and that yields satisfactory predictive performance. Once a feature representation is fixed, institutions adapt their business processes accordingly. For example, when a customer applies for a loan, the bank often automatically estimates the customer's creditworthiness.%
\footnote{\url{https://archive.ph/CTB0f}}
The underlying decision-support system is given input features that have been found to distinguish between borrowers who repaid their loans and those who did not. Since banks need to process loan applications at scale, standardized application forms and workflows to collect these features are set up, so that all applicants are treated the same way.

A key requirement for fairness in algorithmic decision-making is that ``similar people be treated similarly in the classification'' which applies to ``the set of real individuals (rather than the potentially huge set of all possible [representations] of individuals).''~\cite{dwork:2012}. However, there is no single feature set that captures all people uniformly well. To continue the example above, suppose the bank asks loan applicants ``How long have you been at your current job?''. All other things being equal, this feature puts people who recently switched their job at a disadvantage. For this group, the feature is unjustified, even if it may be a good predictor for everyone else. But in practice, system designers and auditors typically assess the trustworthiness of decision-making systems through the outcomes they produce, rather than through the fidelity of their input representations. To the best of our knowledge, no approaches to validate or justify an input representation based on an individual's personal attribute values has yet been proposed (Section~\ref{related_work}).

\begin{figure}[t]
\centering
\includegraphics{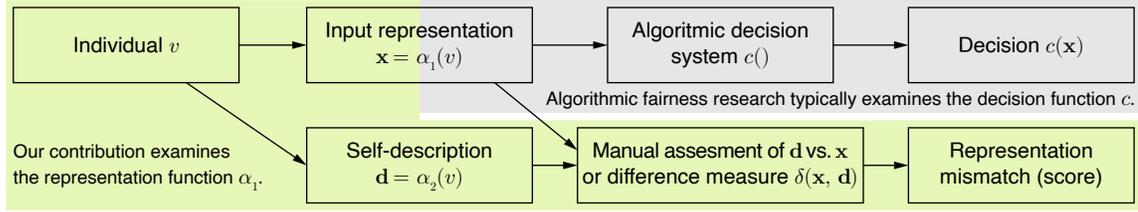}
\caption{Overview of our approach to analyze representation fidelity. Top row: An algorithmic decision system typically implements a standard classification pipeline. An individual~$v$ is represented as~$\mathbf{x}$ using a representation function~$\alpha_1()$. The representation~$\mathbf{x}$ is then fed to the classifier~$c()$ of an algorithmic decision system, which outputs the decision~$c(\mathbf{x})$. Bottom row: Our approach to analyze representation fidelity represents~$v$ as a natural language self-description~$\mathbf{d}$ using a representation function~$\alpha_2()$. The representation~$\mathbf{d}$ is then compared to~$\mathbf{x}$ manually and/or using a distance measure~$\delta()$ to obtain a qualitative and/or a quantitative representation mismatch for the decision~$c(\mathbf{x})$ about~$v$.}
\label{representation-fidelty-analysis-illustration}
\end{figure}

We therefore introduce \emph{representation fidelity} as a new approach to trustworthy computing in algorithmic decision-making (Figure~\ref{representation-fidelty-analysis-illustration}). Our approach allows to systematically analyze if a decision is based on reasonable grounds for the specific individual concerned. When a decision maker cannot offer a satisfying justification for a decision, e.g., when rejecting a loan application from an applicant who recently switched jobs based on ``past job contract duration'', this feature may be inappropriate for deciding about this individual. Our approach is based on comparing, for a given person, two representations: the system's (fixed) input representation and a detailed task-relevant self-description written by the individual who is the decision target. Self-descriptions act as a person-centered representation of independent origin. By relating these two representations, we can measure the extent to which decisions about people can be grounded in their individual characteristics, rather than merely in a population-level aggregate.

Aside from a thorough literature review on related work, our contributions are as follows:
\Ni
we devise the, to our knowledge, first approach to analyze representation fidelity qualitatively, including a typology of representation mismatches, a requirements analysis, and comparative annotation guidelines (Section~\ref{representation-fidelity-analysis}); 
\Nii
we operationalize our approach by compiling the \texttt{Loan-Granting Self-Representations~Corpus~2025}%
\footnote{Data (anonymized for double-blind review): \url{https://doi.org/10.5281/zenodo.17640164}}
dataset to serve as a first large case study and benchmark for evaluating representation fidelity in the context of loan granting (Section~\ref{case-study-loan-granting});
\Niii
we investigate and analyze first baseline measures for quantifying representation fidelity (Section~\ref{quantifying-representational-fidelity}).

%% file: arxiv26-representation-fidelity-part2.tex
\section{Related Work}
\label{related_work}

Our work studies the alignment between human decision targets and their digital representations. This section situates our contribution within abstract discussions of constructs in machine learning, surveys real-world human-targeted decision systems, reviews state-of-the-art approaches to human representation and representational harm, and concludes with prior uses of the German Credit Dataset on which our work builds.
\subsection{Constructs and Observations}

\citet{friedler:2021} propose a structured account of data bias that distinguishes between idealized construct spaces for features and decisions and their imperfect observed counterparts used as proxies in formal decision systems. Reframed in this terminology our work investigates whether a natural language interface can mediate the observational process between the construct feature space and the observed feature space for human decision targets.

\citet{jacobs:2021} further refine the \emph{observational process}~\cite{friedler:2021} through measurement modeling, a social science approach for diagnosing unjustified algorithmic decisions that arise when observable features fail to operationalize task-relevant constructs. This aligns with our premise that decisions cannot be properly justified when observed features do not capture the attributes human decision targets consider relevant. While \citet{jacobs:2021} show that such construct--proxy mismatches lead to systematic errors and inequities, we focus on unjustified decisions by using natural language self-descriptions to surface representation mismatches between construct and observation spaces.

\citet{grgic-hlaca:2018} examine how people assess feature fairness in algorithmic decision-making and find that features perceived as irrelevant or unreliable are judged unfair. For instance, participants viewed the use of high-school grades in recidivism prediction as unjustified because it is unrelated to criminal risk. This highlights the role of perceived relevance in decision justification and directly motivates our approach, which measures misalignment between individuals’ self-described relevant traits and the features used by the algorithm.

\citet{suresh:2021} provide a structured overview of machine learning harms, highlighting \emph{measurement bias} arising from feature and label selection. Building on a construct–observation distinction similar to \citet{friedler:2021}, they identify three sources of measurement bias: construct oversimplification, variation in feature measurement across prediction targets, and variation in measurement accuracy across groups. In our typology of representation mismatches, we also identify feature oversimplification as a key source of harm for human decision targets.

To our knowledge, there are no practical approaches that make human representations open to contestation. While \citet{gebru:2021} propose datasheets with structured questions for dataset creators, our core concern—\emph{how features are selected}—is only implicitly addressed. Their questions focus on data collection for predefined features rather than on feature selection itself, leaving issues of feature completeness and task relevance largely unarticulated.

\subsection{Human-targeted Decision Systems}

Human-targeted decision systems increasingly automate high-stakes decisions in domains such as education~\cite{lamont:2021,perdomo:2025}, employment~\cite{bogen:2019, dastin:2019, fanta:2021}, banking~\cite{martinez:2021, andrews:2021, knight:2019}, or justice and security~\cite{angwin:2016, mak:2021, constantaras:2023, meaker:2023, oosting:2023, brancolini:2022, geiger:2023}. Despite their impact, reports on these systems rarely address the verification and validation of features used to represent human decision targets. Without assessing alignment between individuals and their representations, missing or irrelevant features may go unnoticed, leading to unjustified and difficult-to-contest decisions.

During the COVID pandemic, algorithmic predictions replaced exam-based school grades, relying in part on rankings derived from teachers’ impressions of students’ typical performance~\cite{lamont:2021}. This creates potential mismatches when learning behavior changes, impressions are inaccurate, or rankings depend strongly on individual teachers’ assessments.

\enlargethispage{2\baselineskip}
Another example is the German social benefits system, which determines eligibility for care allowances using a form that rates independence in daily activities across manually curated categories (e.g., mobility, personal hygiene) on a four-point scale. The official documentation~\cite{windeler:2011, bfg:2009} provides limited insight into how these categories were selected. The form inadequately captures episodic conditions, revealing a structural mismatch between standardized task-level features and the complex, fluctuating experiences relevant for eligibility decisions. As a result, assessments of applicants with variable care needs may be unjustified, highlighting the need for more flexible and context-sensitive evaluation.

\subsection{Representations}

\paragraph{Representational Alignment}
The meaning of \emph{representational alignment} differs across the literature: \citet{sucholutsky:2025} survey various understandings in cognitive science, neuroscience, and machine learning, formulating a collective research interest: ``How can we measure the similarity between the representations formed by these diverse systems? Do similarities in representations then translate into similar behavior? If so, then how can a system's representations be modified to better match those of another system?''. Our research relates to the common interest of measuring similarities of representations, but we take a step back, questioning the representation fidelity for human individuals.

\paragraph{Self-Representations}
One approach to representing humans is through self-representation in controlled settings, such as collaborative filtering in recommender systems~\cite{resnick:1994}, where users express preferences by rating items and recommendations are derived from comparisons of these rating profiles. Unlike our setting, collaborative filtering relies on a closed feature set that can represent all users uniformly. In contrast, human representation in decision tasks draws from an open feature set encompassing potentially any individual characteristic.\footnote{While collaborative filtering may incorporate additional personal information, such open-set features remain susceptible to individual misrepresentation.} We therefore employ natural language self-descriptions, which enable nuanced representations and help uncover individually invalid features.

Relatedly, \citet{chandra:2025} present a non-human self-description called a \emph{gauguine}, a probabilistic program that empirically infers its own source code from self-descriptions. The inputs to their program are prior answers to the question ``What am I?'', and the output is a new answer to this question. The gauguine generates multiple different self-descriptions, implementing a suggestion by \citet{lu:2022}. This resembles the social psychology notion of self-concept, which can be described as the answer to the question ``Who am I?''~\cite{myers:2009}.

\paragraph{Representation Learning}
\citet{bengio:2013} surveyed 227~papers on representation learning in~2013. Since then, significantly more work in this area was published based on advances in natural language processing and computer vision~\cite{mikolov:2013, kingma:2022, vaswani:2017, devlin:2019, radford:2021}. Almost all work on representation learning focus on representing objects that are either born-digital or digitized, and on compressing them by selecting the most relevant features from a finite set. Only a small fraction of works consider representations of humans (e.g.,~\cite{athanasiou:2022, zhu:2022, pavlakos:2019}), and they primarily focus on human physiology. To our knowledge, our work is the first to examine the validity of a human's representation for decision tasks.

\citeauthor{bengio:2013} distinguish representation learning from other machine learning tasks in terms of the difficulty of specifying a clear training objective. Representation learning is not primarily about minimizing the number of misclassifications (a common objective in classification). We argue that \emph{representation fidelity}, namely measuring the divergence of a complex real-world entity and its learned representation can serve as a meaningful objective.

\subsection{Mitigating Representational Harms}

Similar to our research, related work on mitigating representational harms is concerned with transparency~\cite{rudin:2019}, mutability~\cite{lu:2022, radlinski:2022}, and fairness~\cite{corvi:2025, katzman:2023, yona:2018, kim:2018, gillen:2018, ilvento:2020, dwork:2012, dwork:2021}. Unlike our research, however, none of these works focus on validating the faithfulness of representations with respect to the human individuals they are intended to represent.

\paragraph{Transparency, Mutability, Fairness}
\citet{rudin:2019} call for prioritizing transparent models over black-box AI in high-stakes decision-making about human individuals. Our paper focuses on detecting representation mismatches, which we view as a step toward greater transparency at the representational level for human classification targets in high-stakes decision tasks.
\citet{lu:2022} appeal to the AI community in general, and the fairness community in particular, to model human identity not statically but as autopoiesis (Greek for ``self-creation''), a theory of identity as a continuous process of interaction with the world rather than a fixed state of mind, so that user representations should also be modeled ``in constant flux.'' In their perspective paper, \citet{radlinski:2022} present such a user representation \emph{in flux} for recommender systems. They envision flexible user representations that allow people to manually modify otherwise automatically generated profiles using natural language descriptions of themselves, and they emphasize the transparency implications of natural language user profiles. Our work aims to adapt and operationalize natural language representations for algorithmic decision-making, taking a step toward \citet{lu:2022}'s notion of autopoiesis.There is a long history of analyzing representational harms through the lens of algorithmic fairness. The topic has gained renewed attention with the emergence of LLMs, with a common goal of identifying representational harms in machine-generated output. In contrast, our work focuses on representations that serve as \emph{inputs} to algorithmic decision systems and considers harms not primarily at the level of protected groups, but at the level of individuals. Despite this difference, \citet{corvi:2025} take an approach that is methodologically similar to ours by proposing a typology of representational harms and developing methods to identify them in generated text. Moreover, \citet{katzman:2023} support our view that self-descriptions (self-identification) are important for reducing representational harms.

Compared to group-based views on representational harms, our work focuses on representational harms to individuals, independent of protected-group membership. Individual fairness, introduced by \citet{dwork:2012}, posits that similar individuals should be treated similarly with respect to a given decision task. Rather than grappling with the potentially large set of possible representations of individuals, \citet{dwork:2012} model individuals as single instances in a metric space. Subsequent research has largely focused on defining, learning, or circumventing the need for a universal distance metric between such individuals~\cite{yona:2018, kim:2018, gillen:2018, ilvento:2020}. Our work expands the notion of an individual by associating each person with alternative representations in the form of natural language texts. We explore how distances between these representations and the original feature representation can be used to validate original feature sets.

\subsection{The German Credit Dataset}

We use the German Credit Dataset~(GCD)~\cite{hofmann:1994} to compile the \texttt{Loan-Granting Self-Representations~Corpus~2025}. We select the GCD as a starting point for our study of representation mismatch detection based on findings from \citet{fabris:2022}. They survey 200~datasets used in algorithmic fairness research and report that, with 35~usages among the surveyed papers, the GCD is the third most frequently used dataset after COMPAS~\cite{propublica:2016} and Adult~\cite{becker:1996}.

Other research using the GCD studies fairness~\cite{pedreschi:2008, ruggieri:2010, kanubala:2024, mehrabi:2022}, data quality~\cite{gromping:2019}, reproducibility~\cite{simson:2025}, and explainability~\cite{dastile:2022}. The dataset is also included in several \texttt{R}~packages, such as \texttt{evtree}, \texttt{CollapseLevels}, \texttt{caret}, \texttt{gamclass}, \texttt{klaR}, \texttt{rchallenge}, and \texttt{scorecard}~\cite{gromping:2019}. We pick the GCD as a starting point for representation mismatch detection because, in contrast to Adult and COMPAS, it comes with a single real-world task and a manageable number of features with a clear encoding. In our view, the downsides of the GCD noted by \citet{fabris:2022} (data from the~1970s; oversampling of bad credits) are less concerning for input fidelity validation than they are for algorithmic credit scoring.

%% file: arxiv26-representation-fidelity-part3.tex
\section{A Qualitative Analysis Approach to Representation Fidelity}
\label{representation-fidelity-analysis}

We develop a new systematic approach for qualitatively evaluating the representation fidelity of human decision targets in algorithmic decision-making (Figure~\ref{representation-fidelty-analysis-illustration}). Let~$V$ denote the set of real human individuals and~$\mathcal{A}$ the set of all possible representation functions that map humans to a specific type of computational representation. The set~$\mathcal{A}$ includes the feature representation function~$\alpha_1: V \to X$ of a given algorithmic decision system, which maps each individual~$v \in V$ to a $p$-dimensional feature representation~$\mathbf{x} \in X$ (typically an inner-product space such as~$\mathbb{R}^p$). The decision support system implements a decision function in the form of a classifier $c: X \to C$, which maps feature representations to a set of classes~$C$; thus, $c(\mathbf{x})$ denotes the system's decision for individual~$v$. Typically, a single feature representation function~$\alpha_1$ is chosen to represent all individuals in~$V$ uniformly within a given decision system.

Our approach examines a given representation function~$\alpha_1$ by comparing it to a second representation function~$\alpha_2: V \to D$ from~$\mathcal{A}$, which maps the set of individuals~$V$ to the set~$D$ of all possible texts describing an individual. Although natural language processing typically treats plain text as a real-world object to be represented for analysis, from the perspective of algorithmic decision-making a text~$\mathbf{d} \in D$ that describes a human individual~$v$ is itself just another representation of~$v$. Nevertheless, natural language text can capture not only facts but also rich semantics and pragmatics, enabling highly individualized representations and interpretations. Indeed, texts already form the basis for many high-stakes decisions about humans in settings where the decision targets are not physically present.

For this reason, we choose self-descriptions provided by the human decision target as a particularly high-fidelity representation and use them as a point of reference. Concretely, for a given individual~$v$ we compare the feature representation $\mathbf{x}=\alpha_1(v)$ to the textual self-description $\mathbf{d}=\alpha_2(v)$, by manual assessment and/or by quantifying their discrepancy with a representation difference measure~$\delta: X \times D \to \mathbb{R}$. The degree to which a decision about an individual~$v$ is based on a representation~$\mathbf{x}$ that deviates from the individual's self-description~$\mathbf{d}$ can thus be interpreted as a measure of representation fidelity.

In what follows, we present the three components of our approach, namely a typology of representation mismatches that adversely affect representation fidelity, a requirements analysis of self-descriptions, and building on our typology, a generic method to manually analyze input fidelity between input representations and corresponding self-descriptions.

\subsection{Typology of Representation Mismatches}
\label{typology}
We present a typology of representation mismatches between individuals' input representations~$\mathbf{x}$ and their self-descriptions~$\mathbf{d}$. We derived this typology from a manual analysis of 90~pairs of self-descriptions and their corresponding input representations from the German Credit Dataset~\cite{hofmann:1994} (see also Appendix Figure~\ref{fig:worked-example} for an example).%
\footnote{The self-descriptions were generated as part of our case study on loan granting (Section~\ref{case-study-loan-granting}) and comprise synthetic self-descriptions produced by six large language models, each of which generated five self-descriptions for each of three input representations.}
Table~\ref{table-representation-mismatch-typology} summarizes the types of representation mismatches we inferred from this analysis.

Our manual analysis is directed, that is we identify differences of information about an individual's characteristics by comparing self-descriptions to input representations, not the other way around. We assume input representations contain ``prior knowledge'' on individuals which can be enriched by information in self-descriptions. Vocabulary and descriptions in the typology follow this intuition with the effect that omitted information can only affect an information unit in the self-description and never in the input-representation. If information is missing from the input representation, but present in the self-description, we classify it as \emph{additional} in the self-description. This accounts for all directional terms in this typology.

\input{table-representation-mismatch-typology}

We first divide semantic units into three types: aspects, specializations/generalizations, and subjects. Aspects are all single units of information that encode one characteristic of an individual. Reoccurring aspects concerning the same topic can be thematically grouped and named when they belong to a subject. Subjects are also directly extracted from the input representations (i.e., each feature key from the input representation is a subject). Specializations are aspects that provide no additional but more detailed information than the corresponding aspect in the input representation; and---vice versa---generalizations are aspects that contain less detailed information than the corresponding aspect in the input representation. These information units can be characterized by three dimensions concerning knowledge inference type (symmetric/asymmetric), comparison strategy (quantitative/qualitative), and characteristics of their information (additive/omissive/semantic):
\begin{itemize}
\item
\textbf{Type 1 (asymmetric/quantitative/additive)}:
Type~1 representation mismatches include additional aspects, specializations, and additional subjects. Information in them is additive, which means, the self-description~$\mathbf{d}$ provides more information than the corresponding information in the input representation~$\mathbf{x}$.
\item 
\textbf{Type 2 (asymmetric/quantitative/omissive)}:
Type~2 representation mismatches include omitted aspects, generalizations, and omitted subjects. Information is omissive if the self-description~$\mathbf{d}$ provides less information than the corresponding information in the input representation~$\mathbf{x}$. 
\item
\textbf{Type 3 (symmetric/qualitative/semantic)}:
Type~3 representation mismatches include inconsistent aspects, inconsistent specializations, inconsistent generalizations, and inconsistent subjects.
\end{itemize}

Both, Type~1 and Type~2 can be compared quantitatively in terms of which representation provides \emph{more} information. Specialization and generalization are special cases in this regard, as they can only be identified by analyzing corresponding information in both the self-description and the input representations. The knowledge that can be inferred from both of them however differs quantitatively (we consider more detailed information, i.e., specializations, to contain more information than the generic counterpart, and generalizations to contain less information vice versa). Type~1 and~2 representation mismatches are asymmetric, which means a piece of information only exists in either the self-description~$\mathbf{d}$ or the input representation~$\mathbf{x}$ and knowledge on this information unit can only be inferred from one representation. Type~3 representation mismatches are symmetric, which means relevant information can be learned from both corresponding information units, the one in the self-description~$\mathbf{d}$ and the one in the input representation~$\mathbf{x}$. Comparisons of corresponding information units in the self-description and input representations are qualitative. Information in Type~3 representation mismatches is characterized as semantic, which means qualitative insights can be drawn from semantic analysis of two corresponding information units. 

\paragraph{Relevance for Representation Fidelity.}
All types of representation mismatches have a different relevance for determining representation fidelity. We assume Types~1 and~3 to be most relevant, and Type~2 less so. While Type~1 certainly encodes a surplus of information in~$\mathbf{d}$ compared to~$\mathbf{x}$, Type~3 \emph{may} contain an information surplus that can only be determined after semantic analysis. Type~2 mismatches provide information about omitted information in~$\mathbf{d}$ compared to~$\mathbf{x}$, which is less relevant for analyzing representation fidelity as omitted information in~$\mathbf{d}$ is still known to the decision system. In contrast to additional information in~$\mathbf{d}$ that may point to task-relevant characteristics with predictive power that were not considered when computing an individual's decision outcome, the omission of information in~$\mathbf{d}$ does not affect the information considered for computing the decision outcome which is based on~$\mathbf{d}$.

\subsection{Requirements for Self-Descriptions}

Semantically, self-descriptions must be task-relevant, i.e., they should convey information that the individual author deems pertinent for making a decision about them with respect to the task. Formally, we expect self-descriptions to be human-interpretable; for example, they should not be provided as an embedding or other lossy encoding of information. We focus on natural-language self-descriptions because they provide a natural medium for humans to articulate arguments, even about complex subjects. However, we consider other formats and modalities, such as individually authored feature lists or speech, to be valid as well. Additional properties we would expect, but cannot guarantee because they depend on the author population and their experiences, include individuality, truthfulness, formality, and diversity across instances.

Self-descriptions used for analysis or auditing purposes can originate from several distinct sources. They may be written directly by individuals about whom decisions are to be made, for example through voluntary participation in a study. Alternatively, self-descriptions may be manually authored for individuals about whom decisions are simulated, such as in pre-deployment auditing settings. A further option consists of third-party accounts, in which external authors describe individuals who are (simulated to be) subject to decision-making. Finally, self-descriptions may be synthetically generated, for instance using large language models, for early sanity checking and rapid prototyping.

Generating self-descriptions is, however, a methodological step in its own right. It can be realized through different procedures and therefore requires its own qualitative evaluation to assess whether it fulfills its intended role. Across all sources, multiple task-relevant text genres are conceivable, including self-descriptions, application letters, expert opinions or assessments, and legal or advocacy-style texts that argue either in favor of or against the individual.

\subsection{A Generic Annotation Method for Representation Mismatch Analysis of Self-Descriptions}
\label{a-generic-method}

Our generic annotation method for representation mismatch analysis of self-descriptions provides a systematic, schema-agnostic approach to auditing representation fidelity by making mismatches explicit, measurable, and analyzable across datasets and algorithmic decision systems.

\paragraph{Analyzing Type~1 mismatches.}
Given a $(\mathbf{d}, \mathbf{x})$~pair, we treat the features of~$\mathbf{x}$ as a lossy but structured reference schema of task-relevant features and~$\mathbf{d}$ as a free-form semantic completion of~$\mathbf{x}$. Our method exhaustively decomposes~$\mathbf{d}$ into semantically coherent information units and systematically maps each unit to the reference schema. Concretely, our method consists of
\Ni
segmenting~$\mathbf{d}$ into meaning-bearing units, and
\Nii
assigning each unit one or more labels from four label types, namely an existing feature label from~$\mathbf{x}$ whenever a semantic correspondence exists; an \emph{aspect} label for minimal meaning-bearing units that have no corresponding feature in~$\mathbf{x}$; a \emph{specialization} label for units that occur in both representations but are more specific in~$\mathbf{d}$ than in~$\mathbf{x}$; and a \emph{new subject} label that assigns a coherent subject absent from~$\mathbf{x}$ to thematically related aspects across multiple $(\mathbf{d}, \mathbf{x})$~pairs. By enforcing full semantic coverage and allowing multiple labels per unit, our annotation avoids assumptions about relevance or predictive importance and instead aims to faithfully capture points of alignment and divergence between the two representations. We assess the reliability of the mapping process despite its inherent subjectivity via inter-annotator agreement among multiple analysts. Based on the resulting label counts, we compute representation fidelity from Type~1 mismatches as follows:
$$
\mbox{Representation Fidelity}
\ =\ 
|\mbox{Additional~$\mathbf{x}$ labels}|
\ +\ 
|\mbox{New subject labels}|
\ +\ 
|\mbox{Aspect labels}|
\ +\ 
|\mbox{Specializations}|.
$$
Future work will include developing normalized measures of representation fidelity based on label counts.

\paragraph{Analyzing Type~2 mismatches.}
As explained above, Type~2 mismatches are omissive and therefore contribute less relevant information to representation fidelity: information omitted from~$\mathbf{d}$ may still be present in~$\mathbf{x}$, on which decisions are based. Nevertheless, for informational purposes, our Type~1 annotation also considers subjects omitted from self-descriptions by comparing the number of features in~$\mathbf{x}$'s feature set to the number of \emph{first occurrences} of those features in~$\mathbf{d}$.

\paragraph{Analyzing Type~3 mismatches.}
Type~3 mismatches require semantic analysis of the corresponding information in $(\mathbf{d}, \mathbf{x})$~pairs. Such analysis presumes domain-expert understanding of the decision task as well as fact-checking to assess the truthfulness of counterfactual information in a $(\mathbf{d}, \mathbf{x})$~pair, or case-specific relevance assessment at the level of the individual. Accordingly, guidance for annotating Type~3 mismatches must be specified in detail for the particular context and domain of the decision task. After careful assessment, Type~3 mismatches can be reclassified as either Type~1 or Type~2 mismatches at the discretion of the expert. Beyond reclassification, domain experts could also assess the relevance of information, which may differ across individuals and subjects and could be incorporated as weights when computing representation fidelity (e.g., counts derived from low-relevance mismatches in a self-description should contribute less than counts derived from high-relevance mismatches). Developing such a domain expert-driven weighting and adjudication procedure is, however, outside the scope of this paper and of our case study.

\subsection{A Worked Example}

\begin{figure}
\includegraphics[width=\columnwidth]{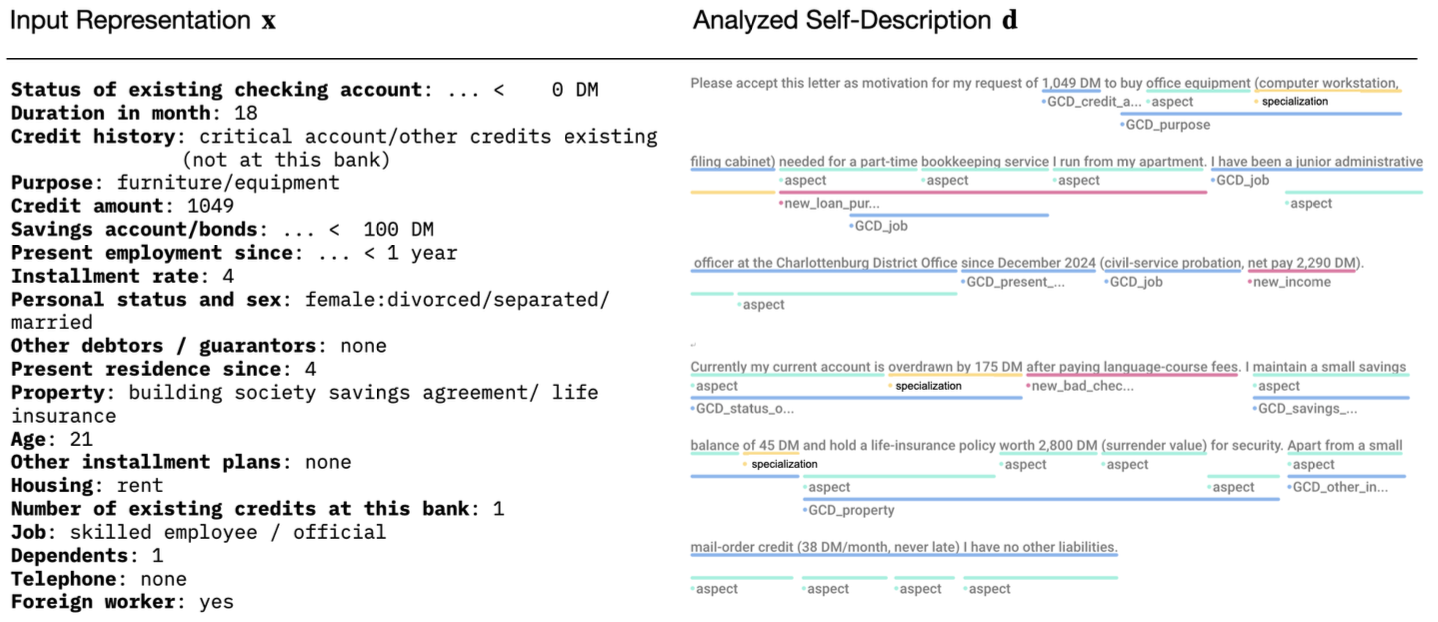}
\caption{This example illustrates our annotation method for representation mismatch analysis by comparing an input representation~$\mathbf{x}$ (left) to a generated natural-language self-description~$\mathbf{d}$ (right). We first segment~$\mathbf{d}$ into information-bearing units, aiming for exhaustive semantic coverage. We then assign to each unit one or more labels from four label types:
\Ni
a label from $\mathbf{x}$'s feature set whenever a semantic correspondence exists,
\Nii
an \emph{aspect} label for any minimal information-bearing unit,
\Niii
a \emph{new subject} label for semantically coherent information units recurring across different self-descriptions, and
\Niv
a \emph{specialization} label whenever a feature in~$\mathbf{x}$ is described in more detail in~$\mathbf{d}$.}
\label{fig:worked-example}
\end{figure}

Figure~\ref{fig:worked-example} illustrates this method with an example of analyzing the representation fidelity of an input representation~$\mathbf{x}$ (left) from the German Credit Dataset using a synthetic natural language self-description generated by gpt-o3~$\mathbf{d}$ (right). Segmenting and annotating~$\mathbf{d}$ into information-bearing units leaves each unit with at least one of the four label types
\Ni
an \emph{$\mathbf{x}$ label} naming scheme in the example ``GCD\_<$\mathbf{x}$-label name>''), that is a label from $\mathbf{x}$'s feature set whenever semantic correspondence is given (blue),
\Nii
an \emph{aspect} label for any minimal information-bearing unit (turquoise),
\Niii
a \emph{new subject} label (naming scheme ``new\_<subject-name>'') for semantically coherent information units across different self-descriptions (pink), and
\Niv
a \emph{specialization} label wherever a feature from the $\mathbf{x}$ is described in more detail (yellow).
After segmentation and annotation are completed, representation fidelity for this example is computed by counting occurrences of all label types and applying the formula presented in Section~\ref{a-generic-method}. As a result, the representation fidelity for the $(\mathbf{d}, \mathbf{x})$~pair in the example is:
$$
\begin{array}{@{}rcl@{}}
\mbox{Representation Fidelity}
&=& 
|\mbox{Additional~$\mathbf{x}$ labels}|
\ +\ 
|\mbox{New subject labels}|
\ +\ 
|\mbox{Aspect labels}|
\ +\ 
|\mbox{Specializations}| \\
&=& 2 + 17 + 3 + 3 = 25
\end{array}
$$

\paragraph{Interpretation.}
To evaluate a decision system's representation fidelity, we compute representation fidelity for each individual~$v$ for which data are available and then summarize the results using descriptive statistics such as the mean and standard deviation. Comparing these statistics across systems for the same decision task enables a relative comparison of representation fidelity across systems. But how should representation fidelity be interpreted at the individual level? The worked example highlights that representation fidelity is inherently approximate. In particular, there is no principled notion of what constitutes a ``good'' level of fidelity, that is, no well-defined bound of the form~$\delta(\mathbf{d},\mathbf{x}) \le \epsilon$. Moreover, the solution space is open-ended: individuals can express arbitrarily many task-relevant aspects in~$\mathbf{d}$, so the absolute value of the measure is not directly interpretable in isolation but only relative to other instances or systems. Nevertheless, comparative statistics can yield meaningful insights. For example, dispersion measures such as the standard deviation can indicate how dissimilar an individual's representations are compared to those of others in the population. Because the annotation task is inherently subjective, representation fidelity estimates may vary across annotators. A practical remedy is to aggregate annotations, for example, via majority voting over labels or by averaging label counts, provided that inter-annotator agreement is sufficiently high.

%% file: table-representation-mismatch-typology.tex
\begin{table*}[h]
\centering
\caption{Differences between individuals' input representations $\mathbf{x}$ and their self-descriptions $\mathbf{d}$ can be classified according to this typology of representation mismatches.}
\renewcommand{\tabcolsep}{11pt}
\begin{tabular}{@{}llll@{}}
\toprule
                                 \multicolumn{4}{c}{\textbf{Types of Representation Mismatches}}                                 \\
\midrule
  No & Name                        & Description                                       & Example                                   \\
     &                             &                                                   & Self-Description vs. Input Representation \\
\midrule
     & Additional Aspect           & aspect in $\mathbf{d}$ absent in $\mathbf{x}$     & ``vegan, no soy'' vs ``diet: no soy''     \\
  1  & Specialization              & more detail in $\mathbf{d}$ than in $\mathbf{x}$  & ``nurse'' vs ``skilled employee''         \\
     & Additional Subject          & subject in $\mathbf{d}$ absent in $\mathbf{x}$    & ``parrot'' vs ``''                        \\
\midrule
     & Omitted Aspect              & aspect in $\mathbf{x}$ absent in $\mathbf{d}$     & ``vegan'' vs ``diet: vegan, no soy''      \\
  2  & Generalization              & more detail in $\mathbf{x}$ than in $\mathbf{d}$  & ``skilled employee'' vs ``nurse''         \\
     & Omitted Subject             & subject in $\mathbf{x}$ absent in $\mathbf{d}$    & ``'' vs ``pet: parrot''                   \\
\midrule
     & Inconsistent Aspect         & meaning in $\mathbf{d}$ differs from $\mathbf{x}$ & ``vegan'' vs ``diet: none''               \\
  3  & Inconsistent Specification  & meaning in $\mathbf{d}$ differs from $\mathbf{x}$ & ``5 years'' vs ``more than 8 years''      \\
     & Inconsistent Generalization & meaning in $\mathbf{d}$ differs from $\mathbf{x}$ & ``more than 8 years'' vs ``5 years''      \\
     & Inconsistent Subject        & meaning in $\mathbf{d}$ differs from $\mathbf{x}$ & ``poodle'' vs ``pet: parrot''             \\
\bottomrule
\end{tabular}
\label{table-representation-mismatch-typology}
\end{table*}

%% file: arxiv26-representation-fidelity-part4.tex
\section{Case Study: Loan Granting}
\label{case-study-loan-granting}

In this section, we present a case study in the context of automated loan-granting decisions that applies the qualitative representation-fidelity analysis approach introduced in Section~\ref{representation-fidelity-analysis}.

\subsection{The Loan-Granting Self-Representations Corpus 2025}
\label{dataset}

We construct the \texttt{Loan-Granting Self-Representations Corpus~2025}, a partly synthetic dataset based on the German Credit Dataset~(GCD)~\cite{hofmann:1994}, by augmenting each input representation of a loan applicant with a generated task-specific self-description, i.e., a loan application letter. For research on representation fidelity, the strengths of this dataset outweigh its limitations (cf.~\cite{fabris:2022}): the~GCD comprises input representations of real-world individuals for a realistic decision task and has an adequate size (1000~individuals).%
\footnote{Code and data: \url{https://anonymous.4open.science/r/individual-representations-E4F9} and \url{https://doi.org/10.5281/zenodo.17640164}}.

\subsection{Generating Natural Language Self-descriptions}

Our corpus of natural-language self-descriptions complements each of the 1000 input representations in the GCD with 30 generated self-descriptions. In this setting, self-descriptions take the form of loan application letters. To ensure letter quality, we specify the following desired properties:
(a)~maintain a formal tone,
(b)~remain mostly coherent with the profile data,%
\footnote{``Mostly coherent'' was sufficient for our use case, as it reflects mistakes in feature sets that also occur in real-world data.}
(c)~include appropriate openings and closings,
(d)~stay focused on the loan request,
(e)~invent plausible specifics for abstract profile attributes%
, and
(f)~differ meaningfully across individuals in content.

\paragraph{Prompt Engineering Strategy}
We iterate over four prompt types on gpt-4.1 (zero-shot, zero-shot+role, zero-shot+role+\allowbreak{}persona, and few-shot+role). For each prompt variant, we evaluate the generated letters with respect to desiderata~(a)--(f) while varying temperature and top-$p$ to identify settings that yield the best overall quality. In a final step, we apply the selected prompt to multiple models (gpt-4.1, gpt-4o, o3, llama-3.3-70b-versatile, moonshotai/kimi-k2-instruct, and qwen/qwen3-32b) and choose the models used to generate the corpus of self-descriptions.
We generate the self-descriptions using zero-shot prompting combined with multiple implicit personas, with temperature~$0.6$ and top-$p{=}0.9$ for all models (except OpenAI's~o3, which does not allow setting temperature or top-$p$). This configuration yields the best results with respect to desiderata~(a)--(f).

\paragraph{Model Selection.} 
A pilot study suggests that the greatest diversity of letters is achieved by combining multiple models. In this study, we manually mapped 50~letters, generated by five models for the same GCD input representation, back to the GCD feature categories. The results indicate that diversity, rather than perfectly detailed or ``likely successful'' letters, is a key asset: our goal is to capture individual variation that reflects real-world decision contexts and supports the analysis of diverse representation mismatches. We observe systematic differences across models in the amount, structure, and repetition of information they produce, with some generating dense, feature-rich letters, while others produce comparatively sparse or repetitive texts. Overall, using multiple models rather than a single model better reflects individual differences in writing quality, feature selection, priorities, style, and even factual correctness, thereby yielding a more realistic and robust dataset. Accordingly, the final corpus contains letters generated by gpt-4.1, gpt-4o, o3, llama-3.3-70b-versatile, moonshotai/kimi-k2-instruct, and qwen/qwen3-32b. Each model was prompted with the same prompt and generated five letters for each of the 1000 input representations from the~GCD.

\subsection{Annotation}
\paragraph{Task.} 
We formulate the annotation problem as a sequence-labeling task over natural language loan application letters generated from the~GCD. The objective is to identify and label all text spans that convey semantically coherent information units and to assign each span one or more subject labels. Our annotation scheme comprises four label types:
\Ni
pre-defined subject labels corresponding to the GCD~input representation, used whenever a span semantically matches an existing GCD~feature;
\Nii
dynamically created subject labels introduced by annotators when relevant information cannot be mapped to any pre-defined GCD~label and reused across the dataset;
\Niii
an \emph{aspect} label, applied to any minimal semantic unit, either as a standalone unit or within a larger coherent span that contains multiple units; and
\Niv
a \emph{specialization} label, marking spans that provide more detailed or fine-grained information than the canonical GCD value templates.

Spans may range from single tokens to entire application letters, and multiple labels may be assigned to a span. The annotation goal is full semantic coverage: every passage that conveys information must be labeled at least once. Annotators are instructed to assign all applicable labels to each span, prioritizing exhaustive semantic matching over task-specific feature selection or assumptions about predictive relevance.

\paragraph{Agreement.} 
Two independent expert annotators labeled 47~distinct application letters sampled from 47~different model outputs. We selected letters to ensure broad coverage of both model diversity and cluster structure. The sample was drawn at random along the distribution of clusters obtained with embedding-atlas~\cite{ren:2025} (Figure~\ref{fig:embedding-atlas-appendix}). First, we drew 40~samples according to the distribution of semantic clusters by selecting one representative per cluster across five rounds and eight clusters. This yielded a balanced mix of models per round (two Kimi, two o3, one Qwen, one gpt-4.1, one gpt-4o, and one Llama), while excluding clusters dominated by Qwen ``thinking'' tokens. Second, we added outlier samples by selecting one outlier per model from each model-specific ``home'' cluster (six in total). As gpt-4o exhibited no meaningful outliers, it was excluded at this stage, whereas two Llama outliers were included due to higher dispersion. In addition, we selected one Qwen sample containing explicit thinking tokens as its own cluster. Finally, we excluded a small cluster of German-language letters (22~from~o3 and 10~from Kimi, covering two profiles) due to its limited size.

We evaluate annotator agreement using two variants of the F1~measure. The first (``strict'') counts a true positive only if both annotators assign the same label to a span with exactly matching start and end offsets. The second (``relaxed'') counts a true positive if both annotators assign the same label and the corresponding spans overlap. Under the strict criterion, agreement is $F1_{\text{strict}} = 0.340$, whereas the relaxed criterion yields $F1_{\text{relaxed}} = 0.765$. We interpret these results as acceptable given the difficulty of the task: annotators had to choose among roughly 30 labels, share fine-grained prior understanding of label semantics, optionally create new labels, and segment and interpret open-ended natural-language text.

\subsection{Corpus Statistics}
Table~\ref{table:dataset-statistics} contains an overview of the number of generated self-descriptions per model and corresponding prompts. Prompts contain input representations from the GCD and corresponding generated self-descriptions are natural language application letters for a bank loan from the point of view of 5~different versions of persons that fit the input representation. Self-descriptions are generated using 6~different large language models (gpt-4.1, gpt-4o, o3, llama-3.3-70b-versatile, kimi-k2-instruct, and qwen3-32b). Each model generated a total of 5\,000~self-descriptions, based on 1\,000~different input representations from the GCD. Additionally, all models produced 5~application letters that are not based a specific input representation from the GCD, but instead the input representation values in the prompt was replaced with the input representation's subject name. In total we obtain 30\,000~self-descriptions that are input-representation value based and 30~"free" self-descriptions which are not. Of those, 47~self-descriptions are annotated by 2~experts.

\begin{table}[t]
\centering
\caption{Overview of generated self-descriptions and prompts. }
\begin{tabular}{@{}lrrrr@{}}
\toprule
  \textbf{Model}          & \textbf{Generated Self-descriptions} &      & \textbf{Prompts} &      \\
                          &                          $\mathbf{x}$-based & free &      $\mathbf{x}$-based & free \\
\hline
  gpt-4.1                 &                               5\,000 &    5 &           1\,000 &    1 \\
  gpt-4o                  &                               5\,000 &    5 &           1\,000 &    1 \\
  o3                      &                               5\,000 &    5 &           1\,000 &    1 \\
  llama-3.3-70b-versatile &                               5\,000 &    5 &           1\,000 &    1 \\
  kimi-k2-instruct        &                               5\,000 &    5 &           1\,000 &    1 \\
  qwen3-32b               &                               5\,000 &    5 &           1\,000 &    1 \\
\hline
  \textbf{Total}          &                              30\,000 &   30 &           1\,000 &    1 \\
  \textbf{$\sum$}         &                              30\,030 &      &           1\,001 &      \\
  \textbf{Annotated}      &                                   47 &      &                  &      \\
\hline
\end{tabular}
\label{table:dataset-statistics}
\end{table}



%% file: arxiv26-representation-fidelity-part5.tex
\section{Quantifying Representation Fidelity} 
\label{quantifying-representational-fidelity}
We quantify representation fidelity using manual methods based on our annotations and we evaluate the usefulness of a first automatic method as baseline to approximate the manual quantification method of representation fidelity.

\subsection{Manual Quantification}

\begin{table}[t]
\centering
\caption{Averaged frequencies of annotated mismatch types per ($\mathbf{x}$-$\mathbf{d}$)-pair.}
\begin{tabular}{@{}ll@{}}
\toprule
  \textbf{Mismatch Type}                                             & \textbf{Average Frequency}  \\
\hline
  (Additional GCD + New Subjects + Aspect Labels) Additional Aspects & (2.11 + 4.78 + 15.32) 22.20 \\
  specializations                                                     & 1.45                        \\
  Omitted Subjects                                                   & 7.07                        \\
\hline
  Representation Fidelity (Additional Aspects + specializations)      & 23.65                       \\
\hline
\end{tabular}
\label{table:dataset-frequencies}
\end{table}

From the annotations, we derive frequencies of mismatch types presented in table~\ref{table:dataset-frequencies} averaged over two annotation results per individual self-description; corresponding individual values are published alongside the code. According to the typology presented in section~\ref{typology}, our annotation scheme allows only for quantification of Type~1 mismatches, because they do not require semantic analysis which can only be provided by domain expert knowledge in loan granting (see section~\ref{typology} for an explanation). 
Averaged additional aspects are composed of the sum of all additional GCD labels, new subjects and aspect labels.
Specializations are counted and averaged as is.
Omitted subjects are are derived from the total amount of GCD labels in the input representations less the number of all single occurrences of GCD labels in the self-descriptions.
We approximate the averaged representation fidelity for the annotated representations via the information surplus the self-descriptions provide over the corresponding input representations adding all occurrences of Type~1 mismatches. The averaged representation fidelity for this dataset is the sum of additional aspects (the sum of additional occurrences of GCD labels, new subjects, and aspect labels) and specializations. All numbers are averaged over the annotations and finally over the number of annotated files.

\subsection{Automatic Quantification}
Our baseline compares GloVe embeddings of both, input representations $\mathbf{x}$ and self-descriptions $\mathbf{d}$. In the following the experiment is explained and evaluated. Table~\ref{table:automatic-results} presents our results as an overview of the automatic quantification methods for representation fidelity an their correlation (Pearson's r) to the manual quantification results for each annotated Type~1 mismatch component representation fidelity~(a), additional GCD~(b), additional subjects~(c), aspect labels~(d), additional aspects~(e), and specializations~(f).

\begin{table}[t]
\centering
\caption{Overview of the automatic quantification methods for representation fidelity an their correlation (Pearson's~$r$) to the manual quantification results for each annotated Type~1 mismatch component representation fidelity~(a), additional GCD~(b), additional subjects~(c), aspect labels~(d), additional aspects~(e), and specializations~(f).}
\begin{tabular}{@{}llrrrrrr@{}}
\toprule
  \textbf{Method}            & \textbf{Metric}       & \multicolumn{6}{@{}c@{}}{\textbf{Pearson's r}} \\
                             &                       &                  (a) &   (b) &   (c) &   (d) &   (e) &   (f) \\
\hline
  GloVe-wiki-50              & Word Mover's Distance &                 0.45 &  0.30 &  0.37 &  0.36 &  0.42 &  0.48 \\
  GloVe-wiki-50-preprocessed & Word Mover's Distance &                 0.30 &  0.11 &  0.19 &  0.28 &  0.28 &  0.31 \\
  GloVe-twitter-200          & Word Mover's Distance &                 0.48 &  0.35 &  0.41 &  0.38 &  0.45 &  0.49 \\
\hline
\end{tabular}
\label{table:automatic-results}
\end{table}

\paragraph{Embedding Baseline.} 
Our embedding baseline represents input representations $\mathbf{x}$ and self-descriptions $\mathbf{d}$ as GloVe embeddings and measures the word mover's distance between them. We compute the word mover's distance (WMD) as implemented in \citet{rehurek:2010} for the annotated set of 47~$\mathbf{x}$-$\mathbf{d}$ pairs of input representations and corresponding self-descriptions. The word mover's distance allows to compare the difference between texts even if they do not share the same words, which is the case for parts of our data.
We implement 3 variations of the embedding baseline: GloVe-wiki-50 uses an embedding model that was trained on a small part of wikipedia, while GloVe-twitter-200 was trained on a larger amount of words from Twitter. Lastly, GloVe-wiki-50-preprocess uses the same GloVe embeddings as GloVe-wiki-50, but preprocesses the natural language text before embedding by removing words from the self-description that exactly match words in the corresponding input representation. The intuition is that with this preprocessing step differences between $\mathbf{x}$-$\mathbf{d}$ pairs become more important and the influence of similarities decreases.

\paragraph{Results.}  Results are presented in table~\ref{table:automatic-results}. Each row shows Pearson's correlation coefficient (r) for one baseline experiment and reflects the linear correlation between the distance computed for each $\mathbf{x}$-$\mathbf{d}$ pair and the manual quantification results presented above. The overall results for the embedding baseline show a weak positive linear correlation between the word mover's distance of GloVe-embedded $\mathbf{x}$-$\mathbf{d}$-pairs and the representation fidelity as per the manual quantification method. Best, though still moderately weak, results are obtained by GloVe-twitter-200, with $r = 0.49$ for the correlation between counted specializations (f) and word mover's distances of $\mathbf{x}$-$\mathbf{d}$ pairs. Lowest results among the embedding baselines yields the GloVe-wiki-50-preprocessed experiment with $r = 0.31$ for its strongest component---again, the counted specializations.
From these results, we learn that the vocabulary the embedding model was trained on, has an impact on the strength of the correlation. Also we learn that removing words from the self-descriptions that match the input representation has a strong negative effect on the results.

%% file: arxiv26-representation-fidelity-sum.tex
\section{Conclusion}
The assumption that human input representations used in algorithmic decision systems faithfully describe all individuals requires validation to ensure justified decisions. We formalize representation fidelity as an auditing method for algorithmic decision systems with human decision targets and operationalize it by comparing distances between input representations and corresponding self-descriptions in a loan-granting context. We introduce a typology of representational mismatches that reduce input fidelity and release a large corpus of synthetic self-descriptions derived from the German Credit dataset. Benchmarking this corpus reveals a weak positive linear correlation (Pearson’s $r=0.5$) between embedding distances and manually identified additional information in self-descriptions.
In the future, developing a reliable automatic method to estimate task-specific additional information in self-descriptions relative to input representations would benefit both individuals subject to algorithmic decisions and deploying institutions. Such methods could improve insight into decision justification at the individual level and enable comparative assessments of representation fidelity across systems, thereby strengthening explainability, user agency, and contestability in algorithmic decision making.

%% file: arxiv26-representation-fidelity-appendix.tex
\section{Appendices}

\subsection{Case Study: Loan Granting}
\begin{figure}[h]
\includegraphics[width=\columnwidth]{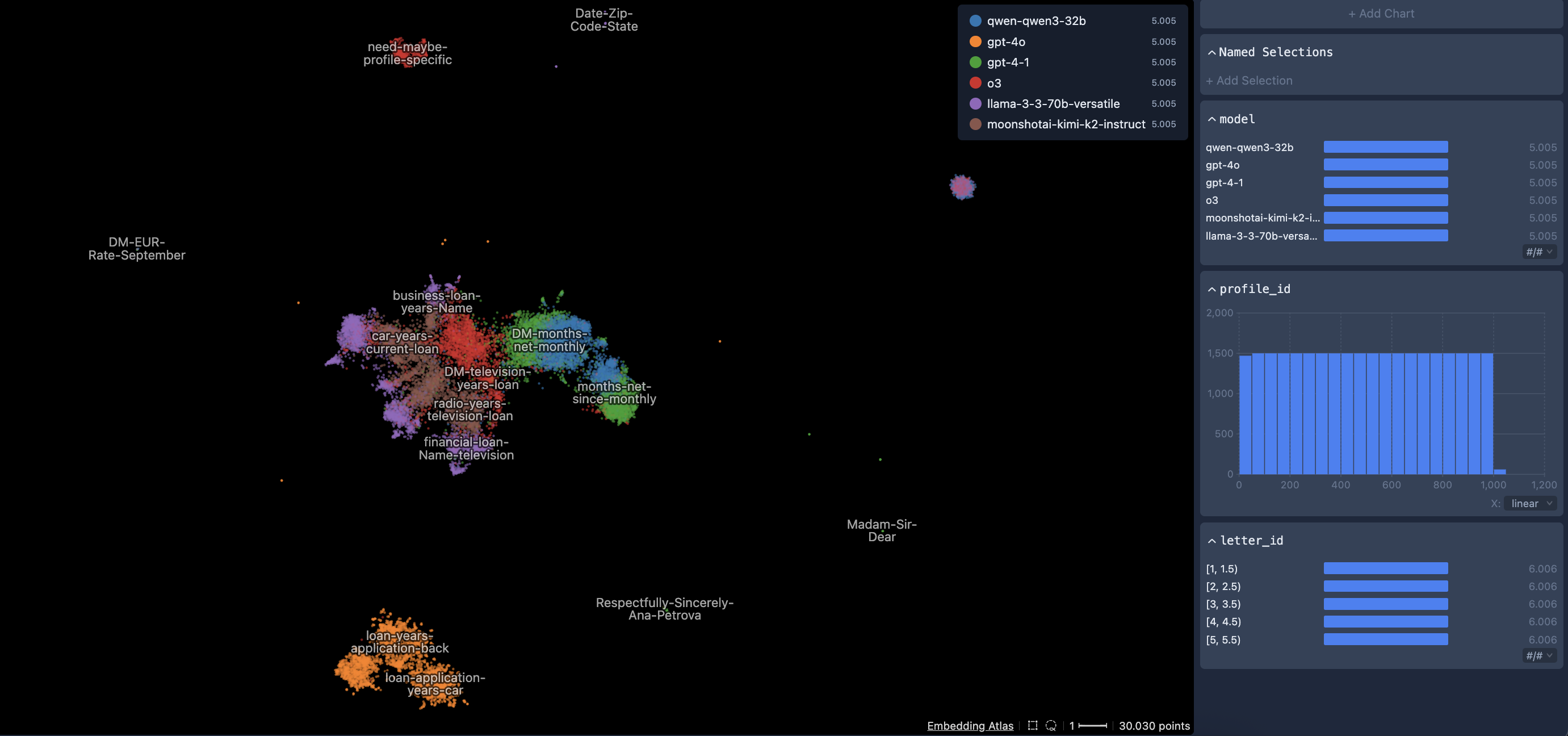}
\caption{This visualization of similarities of embedded self-descriptions as implemented in~\cite{ren:2025}. Self-descriptions form clusters depending the large language model that generated them (gpt-4.1, gpt-4o, o3, llama-3.3-70b-versatile, moonshotai/kimi-k2-instruct, qwen/qwen3-32b).}
\label{fig:embedding-atlas-appendix}
\end{figure}